\documentclass{article} % For LaTeX2e
\usepackage{iclr2026_conference,times}

\usepackage{amsmath,amsfonts,bm}

\def\eqref#1{equation~\ref{#1}}
\def\1{\bm{1}}

\DeclareMathAlphabet{\mathsfit}{\encodingdefault}{\sfdefault}{m}{sl}
\SetMathAlphabet{\mathsfit}{bold}{\encodingdefault}{\sfdefault}{bx}{n}

\usepackage{tcolorbox}
\usepackage{xcolor}

\newenvironment{promptbox}
  {\begin{tcolorbox}[colback=gray!10,colframe=black]}
  {\end{tcolorbox}}

\usepackage{hyperref}
\usepackage{url}
\usepackage{graphicx}
\usepackage{rotating}
\usepackage{booktabs}
\usepackage{wrapfig}

\title{LLM Routing as Reasoning: A MaxSAT View}

\author{Son Nguyen, Xinyuan Liu, Ransalu Senanayake
\\
Arizona State University\\
\texttt{snguye88@asu.edu} \\
}

\iclrfinalcopy % Uncomment for camera-ready version, but NOT for submission.
\begin{document}
\pagestyle{empty}

\maketitle
\begin{abstract}
Routing a query through an appropriate LLM is challenging, particularly when user preferences are expressed in natural language and model attributes are only partially observable. We propose a constraint-based interpretation of language-conditioned LLM routing, formulating it as a weighted MaxSAT/MaxSMT problem in which natural language feedback induces hard and soft constraints over model attributes. Under this view, routing corresponds to selecting models that approximately maximize satisfaction of feedback-conditioned clauses. Empirical analysis on a 25-model benchmark shows that language feedback produces near-feasible recommendation sets, while no-feedback scenarios reveal systematic priors. Our results suggest that LLM routing can be understood as structured constraint optimization under language-conditioned preferences.
\end{abstract}

\section{Introduction}
\paragraph{Routing as reasoning under partial observability.}
LLM routing is the process of dynamically selecting the most appropriate language model for a given query. Most prior work treats routing as learning a policy that trades accuracy for cost. FrugalGPT~\citep{chen2024frugalgpt} learns a cascading strategy for a two-model scenario, routing strong models only when needed. Hybrid LLM \citep{ding2024hybridllm} utilizes the predicted difficulty of a query. Other approaches are trained from preference data~\citep{ong2025routellm}. While effective, these methods predominantly focus on accuracy, and the router typically outputs a single endpoint. With the proliferation of LLMs into real-world industrial use cases \citep{bommasani2022opportunitiesrisksfoundationmodels}, LLM routing has now become an operational problem because a production system must choose among many candidate \emph{LLM endpoints} with different cost--quality trade-offs, and the routing policy is rarely stationary \citep{gama2014drift, Lu_2018}. In addition to non-stationarity, a router faces \emph{partial observability}: endpoints have latent properties (e.g., robustness to long-context errors, instruction adherence, reasoning reliability) that are not fully specified \emph{a priori}. Thus, routing needs to be considered as a \emph{reasoning} problem in which the controller must interpret user intent and act under uncertainty.

\paragraph{Language feedback as constraints.}
A method to route is to provide a short natural-language direction key $d$ (e.g., ``select a cheaper model'') from the current model. The router must map $d$ into a set of constraints over endpoints. Taking a symbolic representation, language feedback (LF) can be injected as logical constraints (hard or soft), and routing corresponds to selecting an assignment that satisfies them.

\paragraph{A MaxSAT view of LLM routing.} Currently, most prior work provides limited support for (i) compositional \emph{natural-language}, and (ii) multi-model scenarios.
In contrast, we formulate routing as a MaxSAT problem (i.e., satisfy as many clauses/constraints as possible)~\citep{ansotegui2013sat, hansen1990algorithms, biere2009handbook}. We propose a unifying abstraction: the router approximately solves a weighted satisfiability instance where (i) \emph{hard clauses} encode safety/feasibility and specification-level requirements, and (ii) \emph{soft clauses} encode preferences whose weights are conditioned on $d$. Importantly, our hard/soft distinction is not the common “hard/soft resource bounds” pattern \citep{chen2026modeling}; rather, \emph{hardness is induced by language semantics} (certainty/negotiability), and weights are \emph{injected per query} by the direction key. Under this view, absence of feedback corresponds to a vacuous constraint set, which should yield a neutral solution unless the router carries implicit priors (i.e., default clause weights).

\paragraph{Contributions.}
\begin{enumerate}
    \item We formalize language-conditioned LLM routing as constrained selection under partial observability, and represent the router's behaviors as logical postconditions.
    \item We provide a reasoning-centric MaxSAT/MaxSMT interpretation in which direction keys inject hard/soft constraints and clause weights, yielding a unified abstraction for both language feedback and default (no-feedback) routing as weighted satisfiability.
    \item We empirically characterize a black-box LLM router on a 25-endpoint benchmark with 34 objective attributes, showing that (i) language feedback produces high-precision, near-feasible shortlists, and (ii) no-feedback routing frequently yields a consistent ``robust core'' explainable by sparse implicit priors $w(\bot)$ rather than unstructured randomness.
\end{enumerate}

\section{Methodology}

\subsection{Problem setup}

\paragraph{Endpoints and partial observability.}
We are given a finite set of $M$ endpoints, $\mathcal{A} = \{a^{(m)}\}_{m=1}^M$. Each endpoint $a^{(m)}$ has
(i) \emph{observable} attributes $\mathbf{o}^{(m)} \in \mathbb{R}^K$ (e.g., output price, cached-input price, context window, max output tokens),
and (ii) \emph{latent} attributes $\mathbf{z}^{(m)}$ (e.g., instruction-following reliability, reasoning stability, failure modes), which are not directly available to the router at decision time.
A routing interaction provides:
$$
(p, \mathcal{A}, a_0, d),
$$
where p is the user prompt, $a_0 \in \mathcal{A}$ is the current endpoint and $d \in \Sigma^\star \cup \{\bot\}$ is a direction key, with $\Sigma^\star$ denoting the set of all finite strings over an alphabet $\Sigma$. We define $d = \bot$ as either not having a language feedback or a direction key that shows no additional preferences (e.g. NONE, I don't have any feedback, etc.)

\subsection{Clause injection}

We consider the router as a single LLM agent, details in Appendix~\ref{app:agent}. The agent receives $(p, \mathcal{A}, a_0, d)$ and outputs a binary output mask C. We model the semantics of a direction key $d$ as a set of constraints over endpoint attributes.
Let $\Phi(d)$ denote a multiset of formulas (constraints) derived from $d$.
In the simplest instantiation, each formula is a predicate over observable attributes,
\[
\varphi(\mathbf{o}^{(m)}, \mathbf{o}^{(0)}) \in \{0,1\},
\]
comparing candidate endpoint $m$ to the current endpoint $0$ (e.g., ``cheaper output price'').

Since natural language is ambiguous, we introduce:
\begin{enumerate}
    \item \textbf{Hard constraints} $\Phi_H(d)$: must-hold requirements (precise keys).
    \item \textbf{Soft constraints} $\Phi_S(d)$: preferences with uncertain scope (vague keys).
\end{enumerate}

We now compile routing into a weighted satisfiability instance.
Introduce decision variables $\{x_m\}_{m=1}^M$, where $x_m=1$ indicates recommending endpoint $a^{(m)}$.

For each constraint $\varphi_j \in \Phi(d)$ and each endpoint $m$, define a satisfaction literal:
\[
\ell_{m,j} := \varphi_j(\mathbf{o}^{(m)}, \mathbf{o}^{(0)}) \in \{0,1\}.
\]

For each soft constraint $\varphi_j \in \Phi_S(d)$ we associate a nonnegative weight $w_j(d) \in \mathbb{R}_{\ge 0}$,
which captures how strongly the direction key $d$ expresses preference $j$. To model routing as selecting a non-trivial shortlist, we impose a recommendation-budget constraint
$L(d) \le \sum_{m=1}^M x_m \le U(d)$ (e.g., $L(d)=1$ under LF).
Moreover, soft constraints should contribute \emph{only when an endpoint is selected}.
We introduce auxiliary variables $y_{m,j} \equiv (x_m \wedge \ell_{m,j})$ and maximize,
\[
\max_{\mathbf{x},\mathbf{y}}
\sum_{m=1}^M\sum_{j\in\Phi_S(d)} w_j(d)\, y_{m,j}
- \lambda(d)\sum_{m=1}^M x_m,
\]
subject to (i) hard constraints $x_m \rightarrow \ell_{m,j}$ for all $j\in\Phi_H(d)$ and
(ii) definitional constraints $y_{m,j}\leftrightarrow(x_m\wedge \ell_{m,j})$.
This yields a weighted MaxSMT instance in which language feedback injects weighted clauses,
while $\lambda(d)$ captures a preference for smaller (more robust) recommendation sets. In our experiments, we do not execute an external solver; instead, we treat the LLM router as a black-box policy and analyze its outputs through this MaxSMT lens by testing whether observed selections are consistent with clause satisfaction and implicit weights.

\paragraph{Regime-specific constraints.}
We distinguish two evaluation modes.
(\textit{i}) \emph{User-facing recommendation} (LF): we impose a shortlist budget $L(d)\le \sum_m x_m \le U(d)$ and use $\lambda(d)>0$ to encourage concise sets.
(\textit{ii}) \emph{Completeness evaluation} (LF-PD): to measure whether the router captures a precise constraint, we disable shortlist regularization by setting $\lambda(d)=0$ and drop the budget constraint, requiring
$x_m \leftrightarrow \bigwedge_{j\in\Phi_H(d)} \ell_{m,j}$ so that all satisfying endpoints are included.

\subsection{Postconditions}

From the binary output mask C, we extract output set $\mathcal{C} \subseteq \mathcal{A}$ with $\mathcal{C} = \{a \in \mathcal{A} \mid C(a) = 1 \}$.
We consider six possible postconditions from the output set:

\begin{enumerate}
    \item Zero($C$) := ($|\mathcal{C}| = 0$)
    \item One($C$) := ($|\mathcal{C}| = 1$)
    \item Some($C$) := ($1 < |\mathcal{C}| < M$, and $\mathcal{C}$ exhibits a \emph{structured} preference pattern)
    \item All($C$) := ($\mathcal{C} = \mathcal{A}$)
    \item Random($C$) := ($1 < |\mathcal{C}| < M$, and $\mathcal{C}$ is \emph{unstructured})
    \item Fail($C$) := $C$ violates the required mask format prior to parsing/normalization
\end{enumerate}

\paragraph{Neutrality as a hard clause.}
When $d=\bot$, the intended postconditions are Zero$(C)$ or All$(C)$. Under no-feedback, neutrality means the distribution over outputs should not systematically privilege particular endpoints beyond an explicitly defined tie-break rule. Any non-trivial subset choice must be explained by \emph{implicit priors}, i.e., default non-uniform weights $w_j(\bot)$ that the LLM router carries even without feedback, forming the postcondition Some(C). This yields a testable prediction: no-feedback routing will preferentially select endpoints that satisfy a stable conjunction of ``robustness'' predicates (e.g., high capability, low price).

\paragraph{Case S.} An output set cannot be determined as Some($C$) or Random($C$) when ($1 \le |\mathcal{C}| \le M$) until we analyze the results. We define this as \textbf{Case S} and our desired result as Case S$^{*}$ $\subseteq$ Case S.
$$\text{Case S} = \text{Some}(C) \vee \text{Random}(C)$$

\paragraph{Failure handling.} To ensure robustness to variations in the LLM agent’s behavior, we require that the output set remain well-defined regardless of the agent’s execution status. In particular, we treat Fail($C$) as a special case of other cases ($|\mathcal{C}| = 0\to$ Fail($C$) = Zero($C$)) $\land$ ($|\mathcal{C}| = 1\to$ Fail($C$) = One($C$)) $\land$ ($1 < |\mathcal{C}| < M\to$ Fail($C$) = Case S).

\paragraph{Processing Rules.} We utilize the following processing rules for $C$:
\begin{enumerate}
    \item Extract all standalone binary tokens {0, 1}
    \item If $|C| < M$, append zeros until $|C| = M$
    \item If $|C| > M$, truncate $|C|$ to its first M elements
\end{enumerate}

\section{Experiments}

\paragraph{Setup.} We evaluate an LLM router over \cite{openai_model_compare} models with $M=25$ endpoints and $K=34$ objective attributes (Appendix~\ref{app:model-zoo}). We remove the model names when feeding the model zoo to the agent to prevent possible external knowledge leakage \citep{Dodge2021DocumentingLW, lee-etal-2022-deduplicating, carlini2023quantifyingmemorizationneurallanguage}.

\paragraph{Direction keys.}
We categorize direction keys into:
(i) \textbf{LF} (language feedback) and
(ii) \textbf{NF} (no feedback; $d\in\{\texttt{NONE}, \texttt{NONE.}\}$).
Within LF, we further distinguish:
\begin{itemize}
    \item \textbf{LF-PD} (precise description): e.g., ``I want a model with cheaper cached input''; ``I want a model with cheaper output prices.''
    \item \textbf{LF-GD} (general direction): e.g., ``I want a cheaper model.''
\end{itemize}
We use output price as the baseline attribute for comparison.

\paragraph{Metrics.}
Let $T(d)$ indicate the \emph{ground-truth target set} induced by an interpreted constraint (e.g., models with strictly cheaper output price than $a_0$).
We evaluate
$
\delta(\mathcal{C};T) = \frac{|\mathcal{C}\cap T|}{|T|}$ and $
\mathcal{P}(\mathcal{C};T) = \frac{|\mathcal{C}\cap T|}{|\mathcal{C}|},
$
with $\delta(\mathcal{C}; T)$ denoting the coverage percentage and $\mathcal{P}(\mathcal{C}; T)$ denoting the precision of the output set.

\begin{table}[ht]
    \centering
    \caption{Count and Percentage of each outcome for different keys. Without specifying \emph{how many models}, the Agent will always recommend some models ($|\mathcal{C}| > 1$) to the user.}
    \begin{tabular}{l|ccccc}
        \toprule
       Direction Key & Zero (\%) & One (\%) & Case S (\%) & All (\%) & Count \\
       \midrule
       I want a model with cheaper cached input.   & 0 & 0 & 12 (100\%) & 0 & 12 \\
       I want a cheaper model.   & 0 & 0 & 5 (100\%) & 0 & 5 \\
       I want a model with cheaper output prices.   & 0 & 0 & 1 (100\%) & 0 & 1 \\
       NONE & 0 & 0 & 101 (51.01\%) & 97 (48.99\%) & 198 \\
       NONE. & 0 & 0 & 10 (50\%) & 10 (50\%) & 20 \\
       \bottomrule
    \end{tabular}
    \label{tab:outcome}
\end{table}

\subsection{LF analysis}

\vspace{-10pt}
\begin{wrapfigure}{r}{0.68\textwidth}
    \centering
    \includegraphics[width=\linewidth]{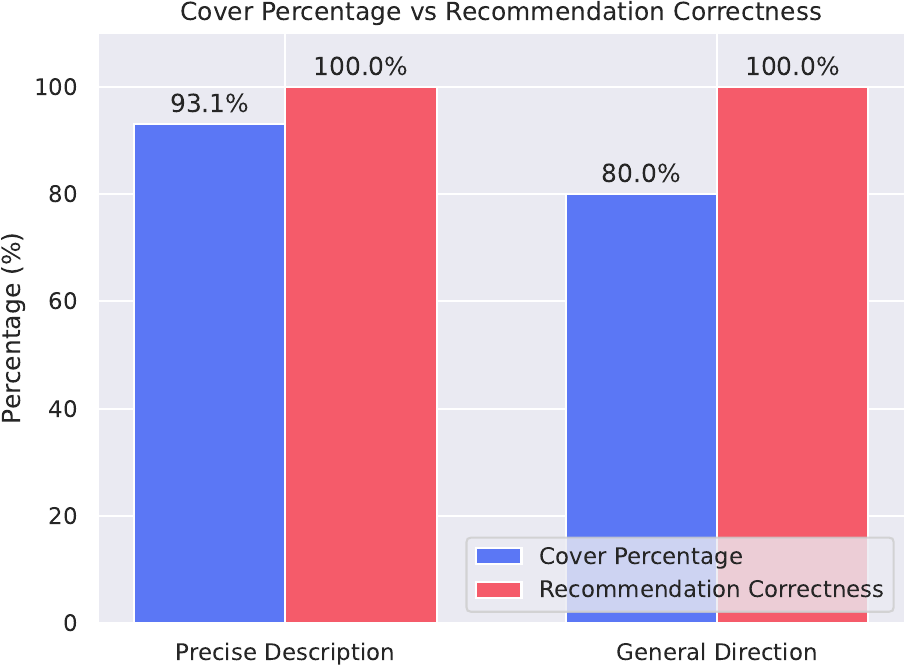}
    \caption{Coverage Percentage and Recommendation Precision of both LF groups}
    \label{fig:cover-percentage}
    \vspace{-20pt}
\end{wrapfigure}
In Figure~\ref{fig:cover-percentage}, the agent achieves $\mathcal{P} = 100\%$ for both scenarios, therefore forming Case S$^{*} = $ Some($C$). However, even when asked to follow a specific direction, the agent still ignores some models ($\delta = 93.1\%$ for LF-PD and 80.0\% for LF-GD). This suggests that under uncertainty, the agent tries to offer a smaller set for robustness. 

\subsection{NF analysis}

\subsubsection{NF results}

When $d=\bot$ (NF), the intended semantics is a vacuous constraint set $\Phi(d)=\emptyset$; thus the neutral postconditions are Zero($C$) or All($C$). However, in around 50\% of cases (Table~\ref{tab:outcome}), its postcondition presents Case S.

\begin{figure}[ht]
    \centering
    \includegraphics[width=1\linewidth]{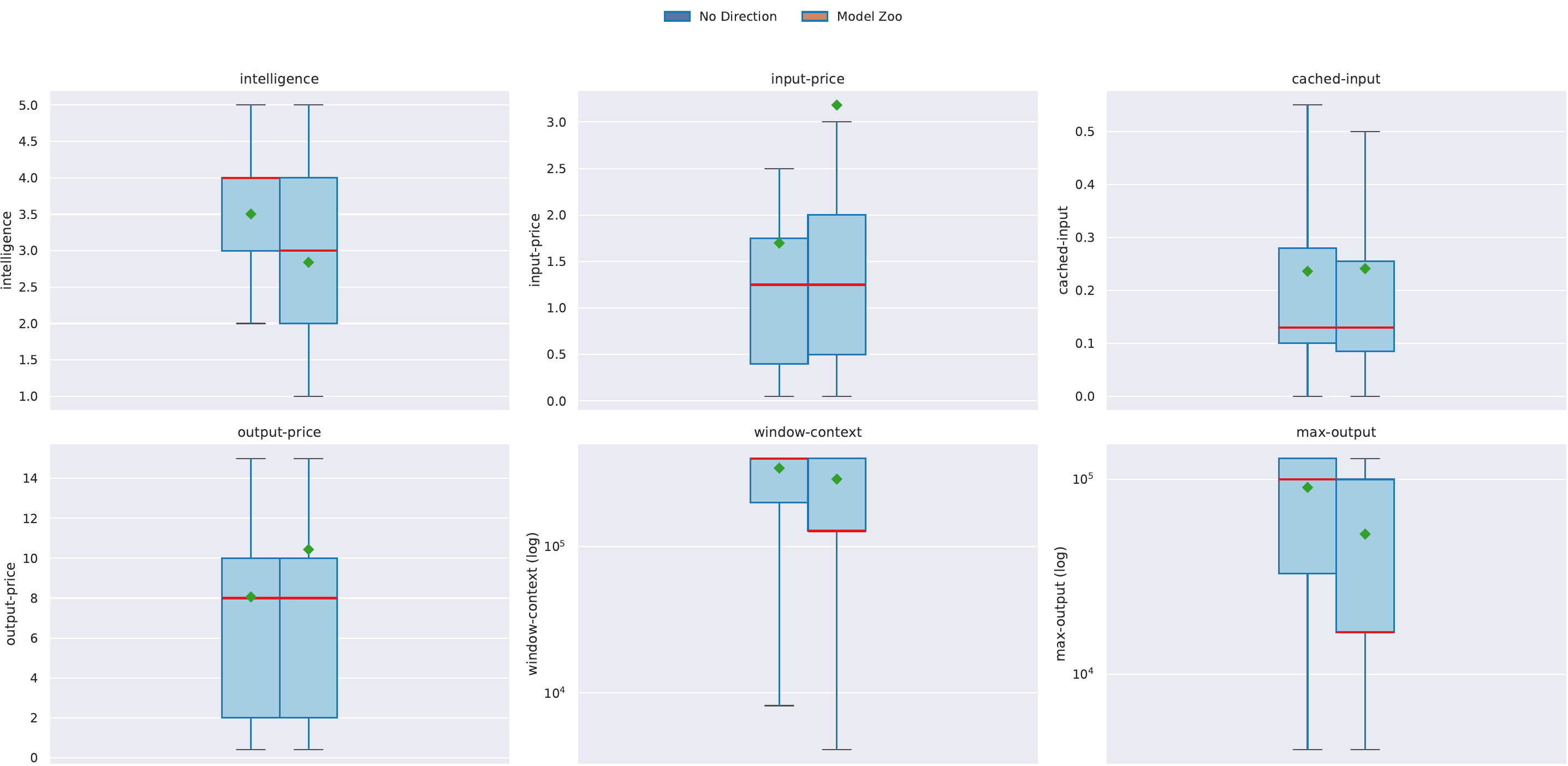}
    \caption{Percentile and Mean of NF set and model zoo on key objectives.}
    \label{fig:ND-attributes}
\end{figure}

\begin{figure}[ht]
    \centering
    \includegraphics[width=1\linewidth]{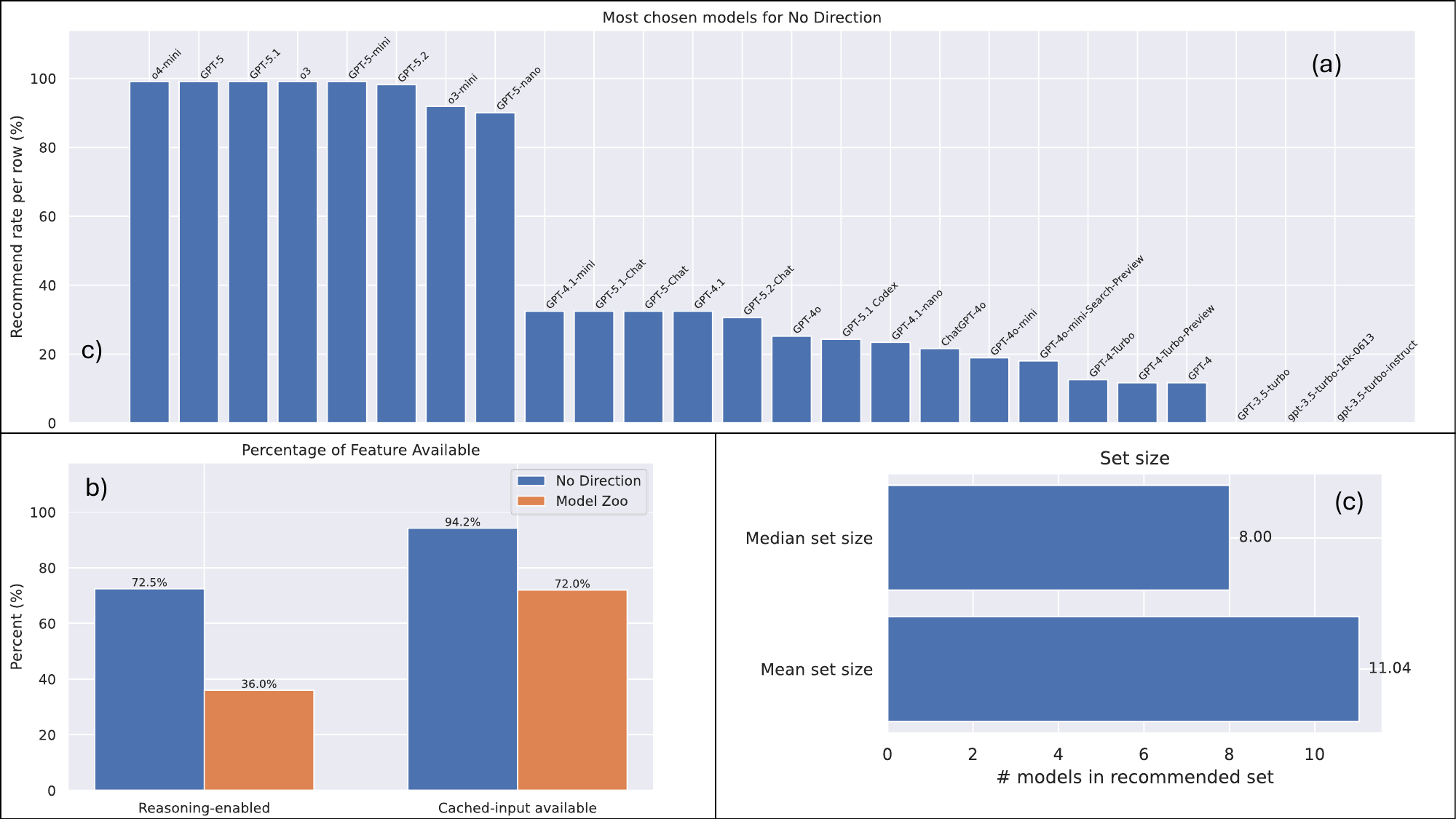}
    \caption{Direction key without language feedback set results for Case S. (a) per-endpoint selection frequency, b) reasoning/cached-input prevalence, (c) set-size statistics.}
    \label{fig:ND}
\end{figure}

As seen in Figure~\ref{fig:ND}, there are eight endpoints with clearly superior recommendation rates: o4-mini, GPT 5, GPT 5.1, o3, GPT 5 mini, and GPT 5.2, o3-mini, and GPT 5 nano. This set of choice has:
\begin{itemize}
    \item The only model with the maximum intelligence score: o3.
    \item A set of intelligence = 4 with lowest price, and the only model with intelligence = 4 and speed $>$ 3.
    \item One endpoint with intelligence = 3 and one with intelligence = 2, both of which have the lowest cost in their respective intelligence groups.
    \item All of them support \emph{cached input}, meaning the potential lower cost at deployment.
\end{itemize}

\textbf{Case S.} The set of recommended endpoints provides a group of robust choices for the user to choose from. The router never recommends endpoints with intelligence $=1$, and the price means are lower than that of the model zoo. Relative to the full model zoo, the sets also exhibit lower mean prices and substantially higher medians of context window and max output.
Furthermore, the majority of recommended endpoints have reasoning enabled and cached input available.

This suggests that the router behaves as if it optimizes a non-uniform default preference structure, i.e., an implicit clause-weight vector $w(\bot)$.

NF behavior is best explained not as arbitrary randomness, but as implicit weighted satisfiability: the router preferentially selects endpoints satisfying a stable conjunction of high-weight ``robustness'' predicates even when the user provides no constraints. We further make the implicit-prior hypothesis falsifiable in the next subsection. This also suggests Case S $=$ Some($C$).

\paragraph{Algorithmic interpretation.}
The fitted $\hat w(\bot)$ induces a per-endpoint prior score
$s_m(\bot)=\sum_j \hat w_j(\bot)\psi_j(a^{(m)})$.
Selecting endpoints with large $s_m(\bot)$ yields an explicit baseline predictor for Some($C$) and makes the MaxSMT abstraction operational: even with $\Phi(\bot)=\emptyset$ at the interface level, the router behaves as if it injects a small set of high-weight robustness clauses by default.

\subsubsection{NF implicit-prior baseline}
\label{sec:wbot}

\paragraph{Data.}
We analyze NF runs ($d=\bot$) with $M=25$ endpoints.
We condition the explanatory model on Case S runs and exclude All($C$) runs, since All($C$) contains no information about relative endpoint preferences.
In our data, out of 218 NF runs we observe 111 Case S runs and 107 All($C$) runs.

\paragraph{Clause library.}
We construct an interpretable predicate library $\{\psi_j\}_{j=1}^J$ with $J=19$ boolean predicates over endpoint metadata.
The full list is:
\texttt{cached\_input\_available, reasoning\_enabled, cached\_and\_reasoning, int\_ge\_4, int\_is\_max, speed\_gt\_3, speed\_ge\_4, cheap\_output\_global\_q0.25, cheap\_input\_global\_q0.25, cheap\_output\_in\_int\_tier\_q0.25, ctx\_ge\_median, ctx\_ge\_p75, maxout\_ge\_median, maxout\_ge\_p75, cheap\_cached\_price\_global\_q0.25, int\_ge\_4\_and\_cached, cached\_and\_cheap\_output\_in\_tier, int\_ge\_4\_and\_cheap\_output\_in\_tier, ctx\_p75\_and\_cached}

Quantiles are computed over the full model zoo.

\paragraph{Model and hyperparameter selection.}
We fit $\ell_1$-regularized logistic regression over run--endpoint inclusion pairs.
To avoid leakage across endpoints within the same run, we select the regularization strength $R$ via group cross-validation where groups are runs.
We scan $R\in\{0.01,0.02,0.05,0.1,0.2,0.5,1,2,5,10\}$ and compute CV-AUC.
We choose the sparsest model whose CV-AUC is within 0.02 of the best CV-AUC, yielding $R=0.02$.

\paragraph{Permutation test.}
To test whether the fitted model exceeds chance while preserving set-size effects, we permute endpoint identities independently within each run (thus preserving $|\mathcal{C}|$ per run), refit the model, and recompute metrics.
$p$-values are computed as the fraction of permuted trials achieving metrics at least as large as the observed metrics.

\paragraph{Explanatory baseline.}
We instantiate this hypothesis by fitting a sparse explanatory model for NF endpoint inclusion conditioned on Case S.
Let $x^{(r)}_m\in\{0,1\}$ denote whether endpoint $m$ is selected in NF run $r$.
We construct a library of $J=19$ interpretable boolean predicates $\{\psi_j\}_{j=1}^J$ over endpoint metadata and fit an $\ell_1$-regularized logistic model:
\[
\Pr(x^{(r)}_m=1 \mid d=\bot,\ \text{Case S})
=
\sigma\!\left(b + \sum_{j=1}^J w_j(\bot)\,\psi_j(a^{(m)})\right),
\]
where $\sigma(\cdot)$ is the logistic sigmoid, $b$ is an intercept, and $w(\bot)\in\mathbb{R}^J$ are clause weights.
We condition on Case S runs because All($C$) provides no information about relative endpoint preference.
Hyperparameters are selected by group cross-validation over runs; we choose the sparsest model within 0.02 AUC of the best cross-validated AUC.

\paragraph{Results.}
Training on the 111 Case S runs, the selected sparse model achieves strong predictive power:
pair-level AUC $=0.8839$ (chance $\approx 0.5$), Spearman correlation between predicted probabilities and empirical per-endpoint selection rates $=0.8810$, and top-8 recovery $=8/8$.
A permutation test that preserves set sizes by permuting endpoint identities within each run confirms statistical significance:
AUC under the null has mean 0.5031 while the observed AUC is 0.8839 ($p=0.005$);
top-8 overlap under the null has mean 2.81 while the observed overlap is 8 ($p=0.005$);
Spearman correlation is also significant ($p=0.020$).
Therefore, Case S exhibits Some($C$) behavior, highly structured and can be captured by a small number of metadata predicates.

\paragraph{Recovered implicit clauses.}
At the selected regularization level, the learned $w(\bot)$ is sparse:
\[
\hat b = -1.8043,\qquad
\hat w(\bot)\neq 0 \text{ on 6/19 clauses.}
\]
Table~\ref{tab:wbot} lists the nonzero weights.
The largest positive weights correspond to reasoning enabled, large max output (maxout $\ge$ 75th percentile), and long context (context $\ge$ median), as well as conjunctions involving cached input and reasoning.
This provides an explicit explanation for the recurring NF robust core: when neutrality is violated (Case S), the router behaves as if it injects a small set of high-weight robustness clauses by default rather than selecting endpoints uniformly at random.

\begin{table}[ht]
\centering
\begin{tabular}{l r}
\hline
Clause $\psi_j$ & Weight $\hat w_j(\bot)$ \\
\hline
$\texttt{maxout\_ge\_p75}$ & 0.9317 \\
$\texttt{reasoning\_enabled}$ & 0.9272 \\
$\texttt{ctx\_ge\_median}$ & 0.8011 \\
$\texttt{cached\_and\_reasoning}$ & 0.7762 \\
$\texttt{ctx\_p75\_and\_cached}$ & 0.2243 \\
$\texttt{cached\_and\_cheap\_output\_in\_tier}$ & 0.0306 \\
\hline
\end{tabular}
\caption{Sparse estimate of implicit NF clause weights $\hat w(\bot)$ fitted on Case S runs (111 runs, $M=25$ endpoints, $J=19$ candidate predicates).}
\label{tab:wbot}
\end{table}

\section{Future Directions.}
In this paper, we showed that language-conditioned LLM routing can be usefully interpreted as structured constraint optimization. While our LF experiments demonstrate high-precision, near-feasible recommendation sets, our analysis primarily focuses on the \emph{no-feedback} (NF) regime, where the intended semantics is a vacuous constraint set. The router frequently violates neutrality by returning a consistent, structured subset rather than All$(C)$ or Zero$(C)$. We view this reproducible structure as evidence of implicit clause weights $w(\bot)$ -- a form of stable, language-free \emph{default reasoning} that can be probed through logical postconditions and explained via a small predicate library, directly connecting routing behavior to LLM logical reasoning and consistency.

The next steps would be to generalize and strengthen our finding by further experimenting on a larger range of agents and prompts. Moreover, we look forward to developing mitigation methods.

\bibliography{lm}
\bibliographystyle{iclr2026_conference}

\newpage

\appendix
\section*{Appendix}

\section{Experiment Setup}

\subsection{LLM agent choice}
\label{app:agent}

We use Grok 4.1 Fast \citep{xai2025grok41}, a model with moderate capabilities in instruction following but fair long context reasoning, achieving 52.7\% and 68.0\%, respectively, from Artificial Analysis benchmarks \citep{ifbench2026}.

%===
\subsection{The "Model Zoo" Dataset}
\label{app:model-zoo}

The 25 OpenAI \citep{openai_model_compare} LLM and all the objective-level metadata are detailed in Table~\ref{tab:openai-model-pool}. When evaluating, we drop \texttt{model} and \texttt{model-id}.

\begin{sidewaystable}
\centering
\tiny 
\setlength{\tabcolsep}{1.5pt}
\renewcommand{\arraystretch}{1.2}
\caption{OpenAI Model Pool}
\label{tab:openai-model-pool}
\newcommand{\rot}[1]{\rotatebox{90}{\raggedleft #1}}

\begin{tabular}{l l l *{34}{c}} 
\toprule
\textbf{id} & 
\textbf{model-id} & 
\textbf{model} & 
\rot{Intelligence} & 
\rot{Speed} & 
\rot{Text In} & 
\rot{Image In} & 
\rot{Voice In} & 
\rot{Video In} & 
\rot{Text Out} & 
\rot{Image Out} & 
\rot{Audio Out} & 
\rot{Video Out} & 
\rot{Reasoning} & 
\rot{Input Price} & 
\rot{Cached Price} & 
\rot{Output Price} & 
\rot{Context Window} & 
\rot{Max Output} & 
\rot{Know. Cutoff} & 
\rot{Comp. Endpt} & 
\rot{Resp. Endpt} & 
\rot{Assist. Endpt} & 
\rot{Batch Endpt} & 
\rot{Fine-Tune Endpt} & 
\rot{Streaming} & 
\rot{Func. Calling} & 
\rot{Struct. Output} & 
\rot{Fine-Tuning} & 
\rot{Distillation} & 
\rot{Pred. Outputs} & 
\rot{Rate Lim (Free)} & 
\rot{Rate Lim (T1)} & 
\rot{Rate Lim (T2)} & 
\rot{Rate Lim (T3)} & 
\rot{Rate Lim (T4)} & 
\rot{Rate Lim (T5)} \\
\midrule
1 & \path{openai/chatgpt-4o-latest} & ChatGPT-4o & 3 & 3 & 1 & 1 & 0 & 0 & 1 & 0 & 0 & 0 & 0 & 5 & - & 15 & 128k & 16k & 23-Oct & 1 & 1 & 0 & 0 & 0 & 1 & 0 & 0 & 0 & 0 & 1 & - & 30k & 450k & 800k & 2M & 30M \\
2 & \path{openai/gpt-4.1} & GPT-4.1 & 4 & 3 & 1 & 1 & 0 & 0 & 1 & 0 & 0 & 0 & 0 & 2 & 0.5 & 8 & 1M & 32k & 24-Jun & 1 & 1 & 1 & 1 & 1 & 1 & 1 & 1 & 1 & 1 & 1 & - & 30k & 450k & 800k & 2M & 30M \\
3 & \path{openai/o4-mini} & o4-mini & 4 & 3 & 1 & 1 & 0 & 0 & 1 & 0 & 0 & 0 & 1 & 1.1 & 0.28 & 4.4 & 200k & 100k & 24-Jun & 1 & 1 & 0 & 1 & 1 & 1 & 1 & 1 & 1 & 0 & 0 & - & 100k & 2M & 4M & 10M & 150M \\
4 & \path{openai/gpt-3.5-turbo-16k} & GPT-3.5-16k & 1 & 2 & 1 & 0 & 0 & 0 & 1 & 0 & 0 & 0 & 0 & 3 & - & 4 & 16k & 4k & 21-Sep & 1 & 1 & 0 & 1 & 0 & 0 & 0 & 0 & 1 & 0 & 0 & - & 200k & 2M & 800k & 10M & 50M \\
5 & \path{gpt-3.5-turbo-instruct} & GPT-3.5-Inst & 1 & 2 & 1 & 0 & 0 & 0 & 1 & 0 & 0 & 0 & 0 & 1.5 & - & 2 & 4k & 4k & 21-Sep & 1 & 1 & 0 & 0 & 0 & 0 & 0 & 0 & 1 & 0 & 0 & - & 200k & 2M & 800k & 10M & 50M \\
6 & \path{openai/gpt-3.5-turbo} & GPT-3.5-Trb & 1 & 2 & 1 & 0 & 0 & 0 & 1 & 0 & 0 & 0 & 0 & 0.5 & - & 1.5 & 16k & 4k & 21-Sep & 1 & 1 & 0 & 1 & 1 & 0 & 0 & 0 & 1 & 0 & 0 & - & 200k & 2M & 4M & 10M & 50M \\
7 & \path{openai/gpt-4-turbo-prev} & GPT-4-Prev & 2 & 3 & 1 & 0 & 0 & 0 & 1 & 0 & 0 & 0 & 0 & 10 & - & 30 & 128k & 4k & 23-Dec & 1 & 1 & 1 & 0 & 0 & 0 & 0 & 0 & 1 & 0 & 0 & - & 30k & 450k & 600k & 800k & 2M \\
8 & \path{openai/gpt-4-turbo} & GPT-4-Turbo & 2 & 3 & 1 & 1 & 0 & 0 & 1 & 0 & 0 & 0 & 0 & 10 & - & 30 & 128k & 4k & 23-Dec & 1 & 1 & 1 & 1 & 0 & 1 & 1 & 0 & 0 & 0 & 0 & - & 30k & 450k & 600k & 800k & 2M \\
9 & \path{openai/gpt-4.1-mini} & GPT-4.1-Mini & 3 & 4 & 1 & 1 & 0 & 0 & 1 & 0 & 0 & 0 & 0 & 0.4 & 0.1 & 1.6 & 1M & 32k & 24-Jun & 1 & 1 & 1 & 1 & 1 & 1 & 1 & 1 & 1 & 0 & 1 & 40k & 200k & 2M & 4M & 10M & 150M \\
10 & \path{openai/gpt-4.1-nano} & GPT-4.1-Nano & 2 & 5 & 1 & 1 & 0 & 0 & 1 & 0 & 0 & 0 & 0 & 0.1 & 0.03 & 0.4 & 1M & 32k & 24-Jun & 1 & 1 & 1 & 1 & 1 & 1 & 1 & 1 & 1 & 0 & 1 & 40k & 200k & 2M & 4M & 10M & 150M \\
11 & \path{openai/gpt-4} & GPT-4 & 2 & 3 & 1 & 0 & 0 & 0 & 1 & 0 & 0 & 0 & 0 & 30 & 0 & 60 & 8k & 8k & 23-Dec & 1 & 1 & 1 & 1 & 1 & 1 & 0 & 0 & 1 & 0 & 0 & - & 10k & 40k & 80k & 300k & 1M \\
12 & \path{openai/gpt-5.1-codex} & GPT-5.1-Code & 4 & 3 & 1 & 1 & 0 & 0 & 1 & 0 & 0 & 0 & 1 & 1.25 & 0.13 & 10 & 400k & 128k & 24-Sep & 0 & 1 & 0 & 0 & 0 & 1 & 1 & 1 & 0 & 0 & 0 & - & 500k & 1M & 2M & 4M & 40M \\
13 & \path{openai/gpt-4o-mini-search} & GPT-4o-Search & 2 & 4 & 1 & 0 & 0 & 0 & 1 & 0 & 0 & 0 & 0 & 0.15 & - & 0.6 & 128k & 16k & 23-Oct & 1 & 0 & 0 & 0 & 0 & 1 & 0 & 1 & 0 & 0 & 0 & 40k & 200k & 2M & 4M & 10M & 150M \\
14 & \path{openai/gpt-4o-mini} & GPT-4o-Mini & 2 & 4 & 1 & 1 & 0 & 0 & 1 & 0 & 0 & 0 & 0 & 0.15 & 0.08 & 0.6 & 128k & 16k & 23-Oct & 1 & 1 & 1 & 1 & 1 & 1 & 1 & 1 & 1 & 0 & 1 & 40k & 200k & 2M & 4M & 10M & 150M \\
15 & \path{openai/gpt-4o} & GPT-4o & 3 & 3 & 1 & 1 & 0 & 0 & 1 & 0 & 0 & 0 & 0 & 2.5 & 1.25 & 10 & 128k & 16k & 23-Oct & 1 & 1 & 1 & 1 & 1 & 1 & 1 & 1 & 1 & 1 & 1 & - & 30k & 450k & 800k & 2M & 30M \\
16 & \path{openai/gpt-5-chat} & GPT-5-Chat & 3 & 3 & 1 & 1 & 0 & 0 & 1 & 0 & 0 & 0 & 0 & 1.25 & 0.13 & 10 & 128k & 16k & 24-Sep & 1 & 1 & 0 & 0 & 0 & 1 & 1 & 1 & 0 & 0 & 0 & - & 30k & 450k & 800k & 2M & 30M \\
17 & \path{openai/gpt-5-mini} & GPT-5-Mini & 3 & 4 & 1 & 1 & 0 & 0 & 1 & 0 & 0 & 0 & 1 & 0.25 & 0.03 & 2 & 400k & 128k & 24-May & 1 & 1 & 0 & 1 & 0 & 1 & 1 & 1 & 0 & 0 & 0 & - & 500k & 2M & 4M & 10M & 180M \\
18 & \path{openai/gpt-5-nano} & GPT-5-Nano & 2 & 5 & 1 & 1 & 0 & 0 & 1 & 0 & 0 & 0 & 1 & 0.05 & 0.01 & 0.4 & 400k & 128k & 24-May & 1 & 1 & 0 & 1 & 0 & 1 & 1 & 1 & 0 & 0 & 0 & - & 200k & 2M & 4M & 10M & 180M \\
19 & \path{openai/gpt-5.1-chat} & GPT-5.1-Chat & 3 & 3 & 1 & 1 & 0 & 0 & 1 & 0 & 0 & 0 & 0 & 1.25 & 0.13 & 10 & 128k & 16k & 24-Sep & 1 & 1 & 0 & 0 & 0 & 1 & 1 & 1 & 0 & 0 & 0 & - & 30k & 450k & 800k & 2M & 30M \\
20 & \path{openai/gpt-5.1} & GPT-5.1 & 4 & 4 & 1 & 1 & 0 & 0 & 1 & 0 & 0 & 0 & 1 & 1.25 & 0.13 & 10 & 400k & 128k & 24-Sep & 1 & 1 & 0 & 0 & 0 & 1 & 1 & 1 & 0 & 1 & 0 & - & 500k & 1M & 2M & 4M & 40M \\
21 & \path{openai/gpt-5} & GPT-5 & 4 & 3 & 1 & 1 & 0 & 0 & 1 & 0 & 0 & 0 & 1 & 1.25 & 0.13 & 10 & 400k & 128k & 24-Sep & 1 & 1 & 0 & 1 & 0 & 1 & 1 & 1 & 0 & 1 & 0 & - & 500k & 1M & 2M & 4M & 40M \\
22 & \path{openai/o3-mini} & o3-mini & 4 & 3 & 1 & 0 & 0 & 0 & 1 & 0 & 0 & 0 & 1 & 1.1 & 0.55 & 4.4 & 200k & 100k & 23-Oct & 1 & 1 & 1 & 1 & 0 & 1 & 1 & 1 & 0 & 0 & 0 & - & 100k & 200k & 4M & 10M & 150M \\
23 & \path{openai/o3} & o3 & 5 & 1 & 1 & 1 & 0 & 0 & 1 & 0 & 0 & 0 & 1 & 2 & 0.5 & 8 & 200k & 100k & 24-Jun & 1 & 1 & 0 & 1 & 0 & 1 & 1 & 1 & 0 & 0 & 0 & - & 30k & 450k & 800k & 2M & 30M \\
24 & \path{openai/gpt-5.2} & GPT-5.2 & 4 & 4 & 1 & 1 & 0 & 0 & 1 & 0 & 0 & 0 & 1 & 1.75 & 0.18 & 14 & 400k & 128k & 25-Aug & 1 & 1 & 0 & 0 & 0 & 1 & 1 & 1 & 0 & 1 & 0 & - & 500k & 1M & 2M & 4M & 40M \\
25 & \path{openai/gpt-5.2-chat} & GPT-5.2-Chat & 3 & 3 & 1 & 1 & 0 & 0 & 1 & 0 & 0 & 0 & 0 & 1.75 & 0.18 & 14 & 128k & 16k & 25-Aug & 1 & 1 & 0 & 0 & 0 & 1 & 1 & 1 & 0 & 0 & 0 & - & 30k & 450k & 800k & 2M & 40M \\
\bottomrule
\end{tabular}
\end{sidewaystable}

\subsection{Prompts}

\begin{promptbox}

IMPORTANT: Your role is to choose which language models align with the user's preferences and DIRECTION.
\newline\newline
You will receive:

- A free-form prompt that may contain a line 'DIRECTION: ...' describing the user's current preference direction (e.g. wanting a cheaper model).

- Metadata for the last chosen model, so you can compare (e.g. find models that are cheaper, faster, etc.).

- A CSV table of all available models. 
\newline\newline
There are [N] models (rows) in the CSV, in fixed order.
\newline\newline
Your only task:

1) For each model (each row), decide if it matches the user's preferences and DIRECTION, possibly relative to the last chosen model.

2) Output exactly [N] integers (space-separated), one per row in the CSV, in the same order. Use '1' if the model is acceptable, '0' if it is not.
   
3) Output no other words.
\newline\newline
Last chosen model (CSV; may be '(none yet'):
[Last Model Metadata]

CSV of ALL models:
[Full Model Zoo CSV]
\end{promptbox}

\end{document}